\documentclass{article}
\usepackage[table]{xcolor}
\usepackage[final]{neurips_data_2024}
\usepackage{caption}
\usepackage{array}
\usepackage{multirow}

\usepackage[utf8]{inputenc} 
\usepackage[T1]{fontenc}    
\usepackage{hyperref}       
\usepackage{url}            
\usepackage{booktabs}       
\usepackage{amsfonts}       
\usepackage{nicefrac}       
\usepackage{microtype}      
\usepackage{xcolor}         
\definecolor{darkgreen}{rgb}{0,0.5,0}

\usepackage[normalem]{ulem}
\usepackage{multirow}
\usepackage{diagbox}
\usepackage{booktabs}
\definecolor{mygreen}{HTML}{2CB600}
\usepackage{subfig}
\usepackage{xspace}
\usepackage{appendix}
\usepackage{graphicx}
\usepackage{epsfig}
\usepackage{amsmath}
\usepackage{pifont}
\usepackage{tabularx}
\usepackage{seqsplit}
\usepackage{nameref}
\usepackage{varioref}
\usepackage{hyperref}
\usepackage{cleveref}
\usepackage{todonotes}
\usepackage{comment}
\usepackage{array}    
\usepackage{prftree}  
\usepackage{tcolorbox}
\usepackage{enumitem}
\usepackage{amssymb}
\usepackage{pifont}
\usepackage[section]{placeins}
\usepackage{wrapfig}
\tcbuselibrary{listings,breakable,skins}

\newcommand{\benchmarkname}[0]{TACT\xspace}
\newcommand{\Hquad}{\hspace{0.5em}}

\definecolor{lightgray2}{gray}{0.95}

\title{TACT: Advancing Complex Aggregative Reasoning\\with Information Extraction Tools}

\author{
    Avi Caciularu$^{\gamma}$\Hquad Alon Jacovi$^{\gamma}$\Hquad Eyal Ben-David$^{\gamma}$\Hquad Sasha Goldshtein$^{\gamma}$ \\
    \textbf{Tal Schuster$^{\delta}$\Hquad Jonathan Herzig$^{\gamma}$\Hquad Gal Elidan$^{\gamma,\eta}$\Hquad Amir Globerson$^{\gamma,\tau}$}\\
    $^{\gamma}$Google Research \hspace{0.1em} $^{\delta}$Google DeepMind \hspace{0.1em} $^{\eta}$The Hebrew University of Jerusalem \hspace{0.1em} $^{\tau}$Tel Aviv University\\
    \texttt{avica@google.com} \\\\
    \texttt{\url{tact-benchmark.github.io}}
}

\begin{document}

\maketitle

\begin{abstract}
Large Language Models (LLMs) often do not perform well on queries that require the aggregation of information across texts. To better evaluate this setting and facilitate modeling efforts, we introduce \benchmarkname---Text And Calculations through Tables, a dataset crafted to evaluate LLMs' reasoning and computational abilities using complex instructions. \benchmarkname contains challenging instructions that demand stitching information scattered across one or more texts, and performing complex integration on this information to generate the answer. We construct this dataset by leveraging an existing dataset of texts and their associated tables. For each such tables, we formulate new queries, and gather their respective answers. We demonstrate that all contemporary LLMs perform poorly on this dataset, achieving an accuracy below 38\%. To pinpoint the difficulties and thoroughly dissect the problem, we analyze model performance across three components: table-generation, Pandas command-generation, and execution. Unexpectedly, we discover that each component presents substantial challenges for current LLMs. These insights lead us to propose a focused modeling framework, which we refer to as \textit{IE as a tool}. Specifically, we propose to add ``tools'' for each of the above steps, and implement each such tool with few-shot prompting. This approach shows an improvement over existing prompting techniques, offering a promising direction for enhancing model capabilities in these tasks.
\end{abstract}
\begin{figure}[htbp]
    \centering
    \includegraphics[width=0.86\textwidth]{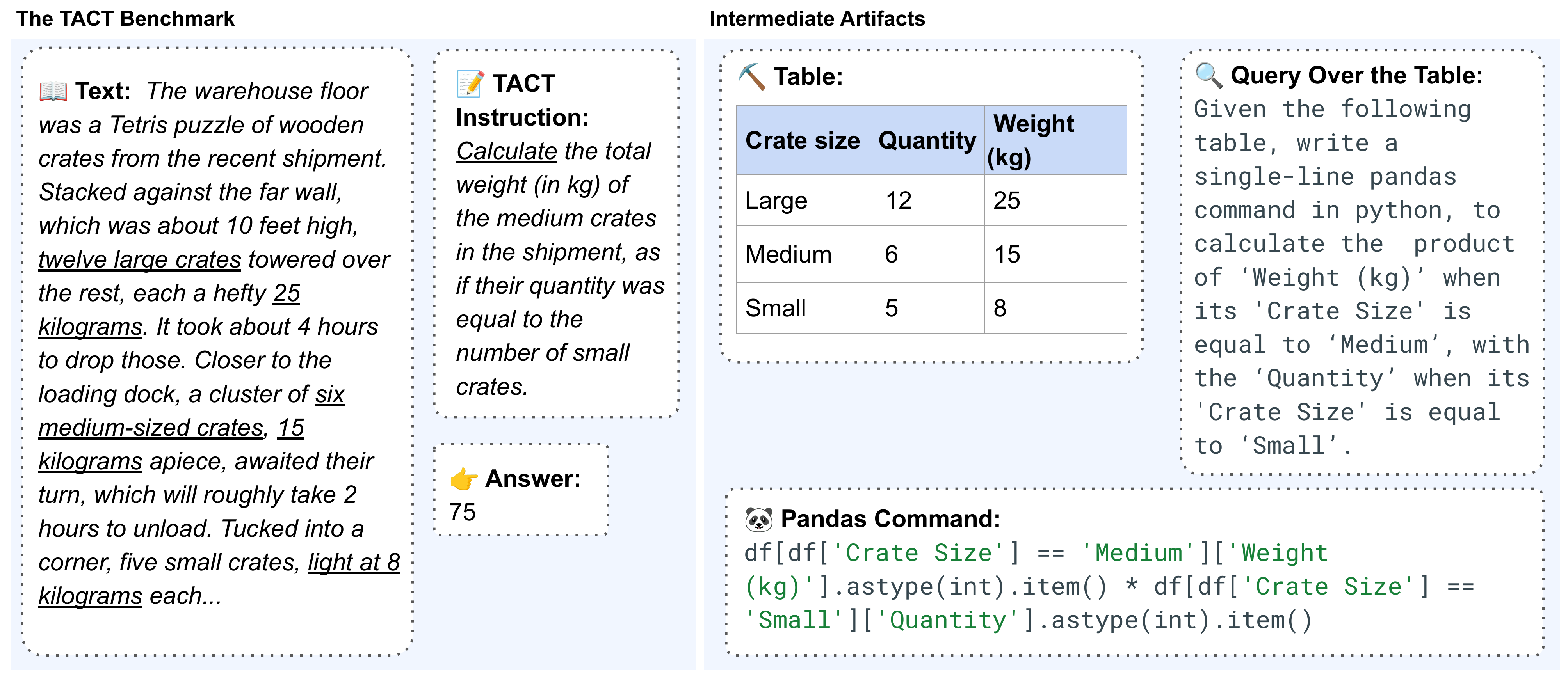}\\
    \caption{Annotated components of the \benchmarkname dataset. The answer is concise but demands advanced reasoning. Intermediate artifacts aid in analyzing LLM reasoning and designing the \textit{IE as a tool} method. Relevant spans are \underline{underlined}.}
    \label{fig:tact_vs_instructie}
\end{figure}
\section{Introduction}
\label{sec:intro}

Large Language Models (LLMs) have shown exceptional capabilities across a wide range of natural language tasks. However, they still face significant challenges in solving complex problems that require reasoning over data presented in non-mathematical formats, such as word and algebraic problems \citep{Amini2019MathQATI,dua-etal-drop-2019}. Interestingly, research indicates that these types of problems pose difficulties not only for LLMs but also for humans \citep{CUMMINS1988405,elliott2023word}. This difficulty mirrors a broader observation about LLM reasoning capabilities: the process of transforming linguistic or graphical inputs into solvable mathematical equations is often more challenging than performing the calculations themselves \citep{schick2023toolformer,das2024mathsensei}. This overarching issue is particularly evident when LLMs attempt tasks that involve the aggregation of information from either single or multiple texts. These models frequently underperform in tasks that require counting, comparing, or processing similar events or entities within texts \citep{caciularu-etal-2022-long,amouyal-etal-2023-qampari,li2023loogle}. This highlights a fundamental limitation: while LLMs can handle isolated data points effectively, their ability to integrate and interpret information across contexts remains a significant hurdle.

A first step towards advancing the capabilities of LLMs on complex reasoning, is to have a high-quality benchmark for evaluating and analyzing their performance in this setting. To this end, we introduce \benchmarkname---Text And Calculations through Tables. \benchmarkname instances were created by NLP and data science experts who wrote aggregative queries over texts. The experts were instructed to use tables (from the InstructIE dataset \citep{Gui2023InstructIEAB}) as the basis for writing the instructions, as they consolidate dispersed information from source texts into a structured format, enabling comprehensive aggregation across the text. The resulting \benchmarkname instances consist of the original text, the written instruction, and a gold answer, all requiring advanced text comprehension and reasoning (illustrated in Figure \ref{fig:tact_vs_instructie}). Importantly, the \benchmarkname task does not include the table, so as to test the ability of the model to answer aggregative queries in an end-to-end manner, requiring advanced text comprehension and reasoning.
Through \benchmarkname, the model is implicitly required to address information extraction (IE) challenges such as coreference resolution \citep{lee-etal-2017-end,joshi-etal-2019-bert,kirstain-etal-2021-coreference}, multi-hop reasoning \citep{lin-etal-2018-multi,dua-etal-drop-2019,Zhao2022MultiHierttNR}, summarization \citep{pmlr-v119-zhang20ae,goyal2022news,slobodkin-etal-2023-dont}, and multi-document processing \citep{caciularu-etal-2021-cdlm-cross,Hirsch2021iFacetSumCI,caciularu-etal-2023-peek,zhu2024fanoutqa}. In Section \ref{sec:dataset}, we describe our methodology for constructing this benchmark---by layering our new expert annotations over InstructIE instances---and the measures we took to ensure its difficulty and robustness.

Having evaluated LLMs on \benchmarkname and observed the challenges it presents to them (Section \ref{sec:tact_exp_and_setup}), we aim to understand the root of these difficulties and explore potential modeling improvements. We propose dissecting the problem into three tasks: table-generation, Pandas command-generation, and command-execution. Leveraging \benchmarkname's ground-truth tables and Pandas commands curated by experts, we analyze the LLM performance of each step in Section \ref{sec:decomposed_exp}. Our findings reveal significant performance headroom in each task, implying that with targeted few-shot prompting, models can considerably enhance their individual task performance. Building on these results, we propose a focused modeling strategy termed the \textit{IE as a tool} framework, which specifically addresses each phase independently (see an illustration in Figure \ref{fig:ie_as_a_tool} and more details in Section \ref{sec:ie_as_a_tool}). This approach has shown to be superior to existing prompting techniques, as detailed in Section \ref{sec:tact_exp_and_setup}. The demonstrated improvements suggest a promising direction for enhancing LLM capabilities in complex reasoning tasks, aligning with our initial findings of untapped potential in each dissected component of the task.

Our contributions are summarized as follows:
\begin{itemize}
\item \benchmarkname: An expert-curated, diverse evaluation dataset that challenges LLMs on following aggregative queries, requiring information extraction and complex reasoning.
\item A rigorous analysis of LLM performance on decomposed \benchmarkname tasks, revealing model strengths and weaknesses in table-generation, Pandas command-generation, and execution.
\item Introduction of the \textit{IE as a tool} framework, leveraging the aforementioned sub-tasks as discrete tools, demonstrating up to 12\% improvement over conventional prompting techniques.
\end{itemize}
\section{Dataset}
\label{sec:dataset}

This section introduces \benchmarkname---Text And Calculations through Tables---a novel challenge set designed to evaluate and improve the capability of LLMs on complex queries that require integration of information. The data is derived from the InstructIE \citep{jiao-etal-2023-instruct} test set using new expert-annotated labels, as described below. In this section, we detail the data labeling methodology employed to create \benchmarkname, highlighting the steps taken to ensure the \textit{reliability} and \textit{validity} of the labeled data. We first introduce the InstructIE benchmark creation methodology (Section \ref{subsec:InstructIE}), then we introduce our \benchmarkname dataset (Section \ref{subsec:tact}), and finally, we explore the properties and conduct an analysis of \benchmarkname (Section \ref{subsec:analysis}). 

\subsection{Background: The InstructIE Dataset}
\label{subsec:InstructIE}
InstructIE \citep{jiao-etal-2023-instruct} is a dataset that includes texts alongside corresponding tables, which summarize the textual content. These tables effectively organize the extracted information into sets of triples—subjects, relations, and objects—derived from the texts. To compile the tables and texts in the test set of InstructIE, which we employed for creating \benchmarkname, human annotators first defined the table topics and columns using real-world texts from the web. These texts were then utilized to craft tables that summarize them through a process combining automatic extraction and manual validation.

The primary components of InstructIE that we utilized are: \textbf{Text}-- the accompanying document or collection of short documents, \textbf{Table}-- a structured representation of the extracted information, where the first row serves as the table header and the subsequent rows contain the extracted data. See an illustrative example in Figure \ref{fig:tact_vs_instructie_both} in Appendix \ref{sec:process}, and \citet{jiao-etal-2023-instruct} for additional details and descriptions of other components that were not included in our study.  While InstructIE provides a good setup for information extraction, it does not directly test the models' abilities to aggregate the extracted information, which we target in \benchmarkname.

\subsection{The \benchmarkname Dataset}
\label{subsec:tact}
Our goal is to evaluate the capabilities of LLMs in addressing aggregative, information-seeking queries, that require both text comprehension and complex reasoning. Using tables as the basis for creating such queries is highly effective, since they consolidate the essential information from their source texts into a structured format. Thus, performing an aggregation on these tables is equivalent to executing an aggregation across the entire text. We leverage and extend the use of the InstructIE dataset, which already contains structured information in table format (see above). We introduce the Text And Calculations through Tables (\benchmarkname) challenge set, aimed at verifying the capabilities of LLMs in handling complex numerical instructions (the resulting \benchmarkname dataset and its components are compared to InstructIE in Figure \ref{fig:tact_vs_instructie_both} in Appendix \ref{sec:process}).

\benchmarkname was created by NLP and data science experts, who employed a rigorous annotation process to transform InstructIE instances into a format suitable for aggregative instruction following. Creating the data includes the steps of assessing the text and the table, then formulating a query in natural language, and finally translating the query into a Pandas command, and executing it on the table.
We chose Pandas over other languages, such as SQL, due to its simplicity. While SQL requires defining a schema, Pandas can easily operate on a single dataframe and often provide solutions with just a single line of code.

Additionally, two human passes were conducted over the dataset, where an expert human validator ensured 100\% accuracy. The expert achieved this level of precision given the lack of ambiguity in the questions, further strengthening the reliability of the data.
See the data creation guidelines, summarization of the annotation process, and more details about the data creation in Appendix \ref{sec:appendix}). The full steps are:

\textbf{Initial Review and Relevance Vetting:} A comprehensive review of the InstructIE dataset,  focusing on texts and tables out of InstructIE's test set containing numerical data. Experts identified tables and text segments where numerical data was present and suitable for quantitative instructions. Tables were vetted for numerical integrity and alignment with the text to ensure data quality. For the remaining examples, the experts were tasked to convert the Markdown-formatted tables from InstructIE into the CSV table format (for convenient parsing into the Pandas dataframe format).

\textbf{Numerical Aspect Identification:} Experts identified \textit{specific numerical aspects} within the text and tables---such as years, currencies, population counts, and temperatures---that enable quantitative operations like counting, calculation, and aggregation. This step identifies which aspect of the table data should be incorporated into the instruction.

\textbf{Natural Language Instruction Formulation:} Based on the identified numerical aspects, experts formulated clear and precise natural language instructions over the text that result in a single numerical value. These instructions targeted the numerical aspects with a focus on aggregation functions like sum, mean, and filtering. 

\textbf{Natural Language Query Over the Table:} After formulating the natural language instructions, experts verbalized them into corresponding natural language queries over the tables (see Figure \ref{fig:tact_vs_instructie}). These queries refined the focus on the numerical data within the table, minimized ambiguity, and helped to prepare the Pandas command.

\textbf{Translation to Pandas Commands and Gold Response Extraction:} Next, experts translated the previous natural language query over the table into a Pandas command. Then, they extracted the gold response by executing the formulated Pandas commands over the tables.

\textbf{Command Execution and Validation:} Finally, the extracted responses were manually verified against the expected outcomes derived from the formulated instructions and texts. This validation step ensured that the results were consistent with the intended instructions and the underlying data from both the text and the tables.

Each instance in the dataset consists of (see illustrative example in Figure \ref{fig:tact_vs_instructie}):
\begin{enumerate}
\item Original Text and Table: Sourced from the InstructIE dataset, these elements contain the foundational data and numerical information relevant to the query. The text provides context, while the table offers structured numerical data aligned with the text content.
\item Natural Language Question: A clearly formulated query in natural language that targets specific numerical aspects identified in the text and table. These questions focus on computational tasks like sum, mean, and filtering to challenge the models' understanding and processing capabilities.
\item Natural Language Query Over the Table: After formulating the natural language question, a corresponding natural language query over the table is developed. This step refines the focus on the numerical data within the table, ensuring that the essential information for the computation is precisely delineated and consistent with the intent of the initial question.
\item Pandas Command: A precise translation of the natural language question into a Pandas command. This command is designed to replicate the expected computational process using the original column names from the table, ensuring the accuracy and consistency of the data manipulation.
\item Expected Result: The correct numerical answer derived from executing the Pandas command, serving as a benchmark to validate the models' responses against the ground truth.
\end{enumerate}

The resulting \benchmarkname dataset contains 124 examples,\footnote{Recent studies, through empirical validation, found that 100 examples are sufficient to conduct a high-quality evaluation of LLMs \citep{liang2023holistic,polo2024tinybenchmarks}.} as well as additional 4 examples that serve as optional few-shot examples for in-context learning.\footnote{These examples are sourced from the validation set of InstructIE, but follow the same process for constructing the \benchmarkname, as elaborated in Section \ref{subsec:tact}.} For evaluating performance on \benchmarkname, we employ exact match for the final answer, since it is a single-span (number). For intermediate steps available in \benchmarkname, such as table- and command-generation, we utilize a both similarity metrics (e.g., ROUGE \citep{lin-2004-rouge}) and execution-based metrics (e.g., accuracy of the generated command's output) as described in Section \ref{sec:decomposed_exp}.

\subsection{Exploring the Numerical Challenges in \benchmarkname}
\label{subsec:analysis}

\begin{figure}[htbp]
    \centering
    \begin{minipage}[b]{0.46\textwidth}
        \includegraphics[width=\textwidth]{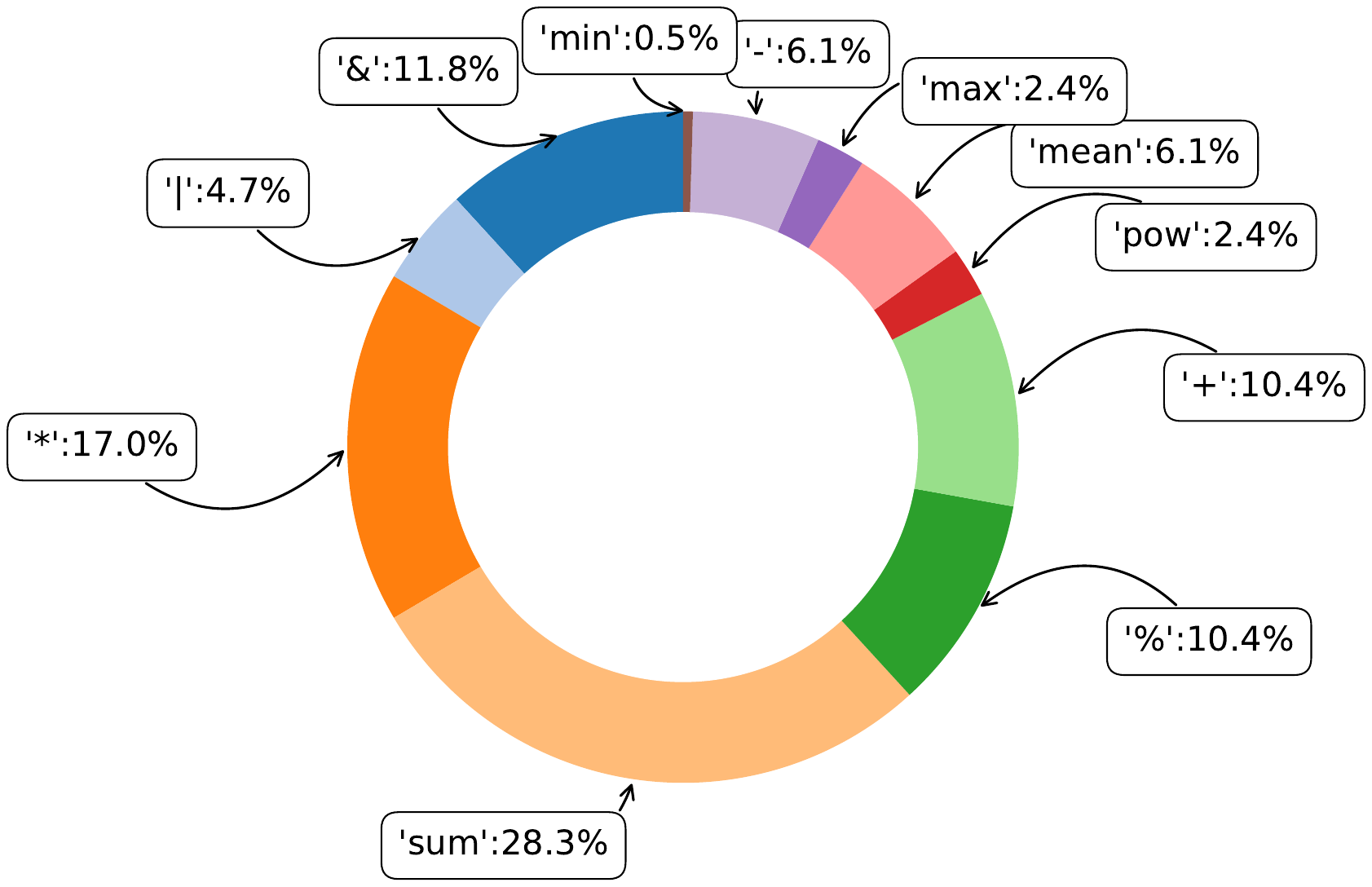}
        \caption{The TACT Dataset pandas different tokens' distribution.}
        \label{fig:operations}
    \end{minipage}
    \hfill
    \begin{minipage}[b]{0.4\textwidth}
        \includegraphics[width=\textwidth]{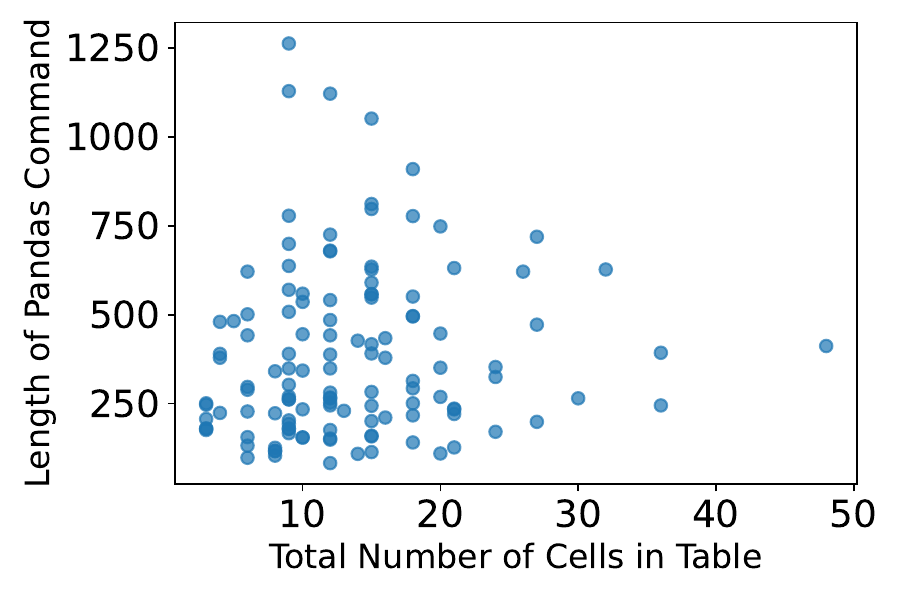}
        \caption{\benchmarkname's Pandas commands' length vs. the total number of cells in their corresponding  tables.}
        \label{fig:tab_stats}
    \end{minipage}
\end{figure}

In this section, we delve into the characteristics of the \benchmarkname dataset. \benchmarkname offers a diverse range of tasks, primarily focusing on two types of instructions--``Calculate'' and ``Count'':

\textbf{``Calculate'' Instructions:} Out of the 124 examples, 63 are categorized under ``Calculate'' instructions. These tasks require the execution of basic mathematical operations to solve the instance. As depicted in Figure \ref{fig:operations}, the operations include addition, subtraction, multiplication, division, and other arithmetic functions. The distribution of these operations, such as summation (28.3\%), mean calculation (10.4\%), and power functions (6.1\%), highlights the varied complexity and the need for precise computational understanding by the models.

\textbf{``Count'' Instructions:} The remaining 61 examples fall under ``Count'' instructions, where the primary objective is to identify specific types or categories within the attached text and perform a simple counting operation. This task challenges the model's ability to accurately parse and interpret textual data, identifying relevant entities or events, and perform the proper counting.

The composition of the \benchmarkname dataset, with a balanced mix of ``Calculate'' and ``Count'' instructions, ensures a comprehensive evaluation of models across different dimensions of numerical reasoning. The operations, as detailed in the pie chart (Figure \ref{fig:operations}), further emphasize the diversity and scope of numerical challenges that \benchmarkname presents, offering a broad testbed for evaluating both fundamental and complex computational reasoning. This variety plays a pivotal role in assessing how well models can generalize their mathematical skills to real-world tasks.

In Figure \ref{fig:tab_stats}, we present a comparison of the total number of cells in a table against the lengths of the corresponding Pandas commands. The figure reveals a wide distribution of data points, illustrating that the length of Pandas commands, quantified in Gemini tokens \citep{team2023gemini}, does not correlate directly with the total number of cells in the tables. The varied spread of points across the graph indicates that additional factors, such as the complexity of arithmetic operations required or the specific data arrangement within the tables, might play a more significant role in determining the length of the commands than simply the volume of data.

In Table \ref{tab:task_types}, we present an illustration of the diverse range of implicit tasks incorporated within TACT, which are specifically designed to test advanced text comprehension and numerical reasoning. Each task is tied to text spans that underline the specific data points or contextual clues necessary for task completion, ranging from multi-document summarization to date and time numerical reasoning. For example, one task leverages coreference resolution, requiring the model to understand and connect information spread across different parts of the text. Another task tests the model's capability for lexical matching, identifying specific words within a context. Complex arithmetic operations are also present, demanding a high level of numerical literacy to interpret numerical and financial concepts. This highlights the interplay between linguistic understanding and numerical computations, demonstrating the ability to handle a wide spectrum of real-world tasks—from simple counting to complex, multi-step mathematical operations embedded within textual data.

\renewcommand{\arraystretch}{1.4} 
\captionsetup[table]{position=below}
\begin{table}[htbp]
  \tiny
  \centering
  \caption{An overview of implicit tasks in TACT, split by their types of 'Count' and 'Calc.' (Calculate) instruction types, along with textual instructions, accompanying texts and their corresponding sub-tasks. The sub-tasks' related aspects in the text are \underline{underlined}.}
  \label{tab:task_types}
  \rowcolors{2}{white}{lightgray2}  
  \resizebox{\textwidth}{!}{
    \begin{tabular}{cc p{0.3\textwidth} p{0.3\textwidth} p{0.15\textwidth}}  
      \toprule
      \multicolumn{2}{c}{\textbf{Operation}} & \multirow{2}{*}{\textbf{Instruction}} & \multirow{2}{*}{\textbf{Relevant text spans}} & \multirow{2}{*}{\textbf{Implicit task}} \\ 
      \cmidrule{1-2}  
      \textbf{Count} & \textbf{Calc.} & & & \\
      \midrule
      & \checkmark & \textit{Calculate the sum of the years that "To Kill a Mockingbird" was published in, and the year that it won a prize according to the text.}
       & \textit{$\ldots$ \underline{It} was published in \underline{1960} $\ldots$ A year after \underline{its} release, \underline{it} won the Pulitzer Prize $\ldots$} & Coreference resolution \\
      \checkmark & & \textit{Count the number of achievements that include instructions.} & \textit{\underline{1.} We present FLAIR $\ldots$ \underline{2.} How well can NLP models perform? $\ldots$ \underline{3.} Pretrained language models have become increasingly prominent $\ldots$} & Multi-document summarization\\
      & \checkmark & \textit{Calculate the sum of squares of the stock price increases in the text.} & \textit{$\ldots$ The S\&P 500 \underline{rose 1.45\%} $\ldots$ the Nasdaq Composite \underline{popped 1.07\%} $\ldots$ The Dow Jones Industrial Average led gains, \underline{rising 2.12\%} $\ldots$} & Complex arithmetics \\
      \checkmark & & \textit{Count the number of weather forecasts that include temperatures between 50 and 91 degrees.} & \textit{$\ldots$ with temps currently ranging from the \underline{upper 80s to low 90s} $\ldots$ Overnight, seasonal temps in the \underline{upper 60s to low 70s} continue $\ldots$} & Numerical range entailment \\
      \checkmark & & \textit{Count the number of wars in the text that have "Indian" within their names.} & \textit{$\ldots$ The American \underline{Indian Wars}, also known as the $\ldots$ and the \underline{Indian Wars} $\ldots$} & Lexical matching \\
      \checkmark & & \textit{Count the number of cases where the delivery date was later than June 2, 2023 and the travel time was more than 2 hours in this text.} & \textit{$\ldots$ 2. On \underline{June 10, 2023}, XYZ Shipping's Truck 789, $\ldots$ Departing at \underline{8:30 AM} and arriving at \underline{2:00 PM} $\ldots$} & Date and time numerical reasoning \\
      \bottomrule
    \end{tabular}
  }
\end{table}
\section{IE as a Tool}
\label{sec:ie_as_a_tool}
\begin{figure}[htbp]
    \centering
    \includegraphics[width=1\textwidth]{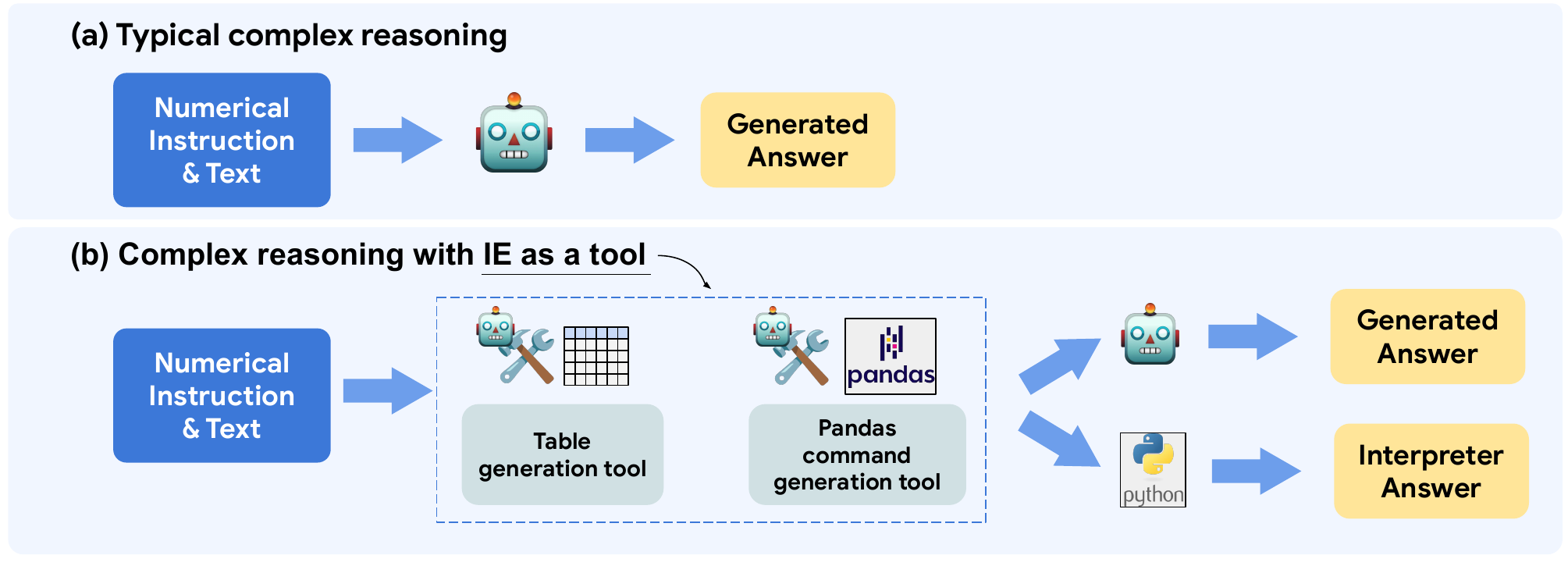}\\
    \caption{Possible setups for solving \benchmarkname with LLMs. (a) Typical Approach: the large language model (LLM) directly generates the answer based on the provided query and text, but without the aid of any external tools. (b) With \textit{IE as a Tool}: This approach utilizes a three-step process. First, an information extraction tool generates a structured table from the text and query. Next, another tool formulates an appropriate Pandas command based on this table, the text and the query. Finally, all this information is fed into the LLM, which then generates the answer; or into a code interpreter that can run the Pandas command over the table.}
    \label{fig:ie_as_a_tool}
\end{figure}

As demonstrated by our experiments in the subsequent sections, current models face significant challenges when tackling the \benchmarkname task. To address this, we introduce a novel approach called \textit{IE as a Tool}, which is illustrated in Figure \ref{fig:ie_as_a_tool}. The core idea is to handle \benchmarkname instructions through the sequential use of two distinct tools: one that generates a table from the provided text and instruction, and another that formulates the corresponding Pandas command. The model then executes the command, alongside the original instruction and text, to derive the final answer. This sequence offers a natural and efficient strategy for addressing \benchmarkname's aggregative queries.

The implementation of these tools can follow multiple methods. For simplicity, we adopted a few-shot prompting approach, as detailed in the prompt templates in Appendix \ref{sec:appendix_prompts}. Our experimental results reveal that \textit{IE as a Tool} yields up to 12\% improvement in performance on \benchmarkname, outperforming conventional prompting techniques (see Section \ref{sec:tact_exp_and_setup}).


\section{Experimental Setup and \benchmarkname Results}
\label{sec:tact_exp_and_setup}

We assess the performance of several models on the \benchmarkname dataset, including GPT-4o \citep{openai2024gpt4}, Claude 3.5 Sonnet \citep{anthropic_claude_2024}, Gemini-1.0-Ultra \citep{team2023gemini}, Gemini-1.5-Pro \citep{reid2024gemini}, Llama-2-13b-chat, Llama-2-70b-chat \citep{touvron2023llama}, Gemma-7b-it (v1.1) \citep{team2024gemma}, Mistral-7b-instruct (v0.2) \citep{jiang2023mistral}, and Mixtral 8x7B \citep{jiang2024mixtral}. Their capabilities are tested using both standard prompting techniques and our \textit{IE as a tool} method (detailed in Section \ref{sec:ie_as_a_tool}). Subsequent analysis focuses on the specific sub-tasks of table generation (Section \ref{subsec:table_gen}) and Pandas command generation (Section \ref{subsec:pandas_gen}), providing a comprehensive evaluation of each model's performance and identifying potential areas for improvement in each sub-task.

We measure performance by averaging over four combinations of few-shot examples (by sampling the order and examples), following the procedure described in \citet{jacovi-etal-2023-comprehensive}, and report the results for \{0, 2, 4\}-shot, using dedicated examples from the validation set of InstructIE (see more details in Appendix \ref{sec:appendix_exp}).

We evaluate the LLMs' performance on the \benchmarkname task, measuring exact match, where the models are provided a numerical instruction and a text, and are tasked to produce the correct numerical answer. We report 4-shot results for Gemma-7b-it, Mistral-7b-instruct, and Mixtral 8x7B, while zero-shot results for the remaining larger models, given the results on table and Pandas query generation in the following sections, and the observation that few-shot demonstrations yield saturated performance for most of the tasks for the larger models. We propose the following experimental setups for each LLM:
\begin{itemize}
    \item \textbf{Generic:} The baseline setting where the LLM receives a \benchmarkname instruction and a text passage, and is tasked with directly generating the answer.
    \item \textbf{Chain-of-Thought (CoT):} This setting is akin to the baseline, with the enhancement of adding ``Let's think step-by-step'' to the prefix (input). This encourages the model to generate a detailed, step-by-step CoT reasoning before producing the final answer.
    \item \textbf{In-context IE:} This method adopts a chain-of-thought-like approach, where the LLM first generates a table from the text, then creates a Pandas query, and finally, with all this context provided, the model generates the answer, all within the same prompt.
    \item \textbf{IE as a tool:} Here, the model can utilize generated tables and Pandas queries to answer the instruction, like the previous setup, however, in this variation, we employ few-shot prompted LLMs as separate tools (the number of shots applies for these tools). See Section \ref{sec:ie_as_a_tool} for more details. We include both \emph{Without Pandas} and \emph{With Pandas} variants as an ablation. We include these since as detailed in Section \ref{subsec:pandas_gen}, even with syntactic errors in the Pandas commands, these errors may still assist the model in generating correct outputs.
    \item \textbf{IE as a tool (Gold):} This configuration is similar to the previous one (with Pandas command) but utilizes gold-standard (i.e., ground truth) tables (Gold Table) and/or Pandas commands (Gold Table+Pandas) from \benchmarkname, rather than relying on outputs generated by the tools. This baseline serves as an upper bound for the performance of the \textit{IE as a tool} approach.
\end{itemize}

\captionsetup[table]{position=below}
\renewcommand{\arraystretch}{1.2} 
\setlength{\tabcolsep}{5pt} 
\begin{table}[htbp]
    \centering
    \tiny
    \caption{Exact match accuracy evaluation results of different models on \benchmarkname, evaluated across different experimental setups, including Generic, Chain-of-Thought (CoT), In-context IE, and IE as a tool with various settings. The best-performing results are highlighted in \textbf{bold}.}
    \label{tab:tact_res}
    \resizebox{\textwidth}{!}{
    \begin{tabular}{lcccccc}
        \toprule
        \multirow{2}{*}{\textbf{Model}} & \multicolumn{1}{c}{\textbf{Generic}} & \multicolumn{1}{c}{\textbf{CoT}} & \multicolumn{1}{c}{\textbf{In-context IE}} & \multicolumn{3}{c}{\textbf{IE as a Tool}} \\
        \cmidrule(lr){5-7}
         & & & & \multicolumn{1}{c}{\textbf{Without/With Pandas}} & \textbf{Gold Table} & \textbf{Gold Table + Pandas} \\
        \midrule
        Gemma-7b (4-shot) & 17.1 & 25.4 & 26.2 & 27.1 / 28.9 & 33.3 & 45.1 \\
        Mistral-7b (4-shot) & 2.4 & 2.6 & 2.6 & 2.7 / 3.5 & 4.1 & 10.9 \\
        Llama-2-13b (4-shot) & 8.5 & 8.0 & 8.3 & 8.5 / 8.5 & 9.2 & 14.3 \\
        Mixtral 8x7B (4-shot) & 4.4 & 4.2 & 5.2 & 5.9 / 6.1 & 6.3 & 12.6 \\
        Llama-2-70b (0-shot) & 3.6 & 4.0 & 4.2 & 4.5 / 7.6 & 12.9 & 22.4 \\
        Gemini-Pro (0-shot) & 28.4 & 34.7 & 12.3 & 40.2 / 41.9 & 48.5 & 72.1 \\
        Gemini-Ultra (0-shot) & 25.4 & 37.3 & 36.6 & 39.7 / 41.4 & 49.8 & 72.3 \\
        GPT-4o (0-shot) & \textbf{30.1 }& 37.7 & 36.4 & \textbf{41.1} / \textbf{42.2} & 50.9 & 74.1 \\
        Claude 3.5 Sonnet (0-shot) & 28.6 & \textbf{37.9} & \textbf{36.8} & 40.8 / 42.1 & \textbf{51.3} & \textbf{74.6} \\

        \bottomrule
    \end{tabular}
    }
\end{table}

The results, which are depicted in Table \ref{tab:tact_res}, show that all models experience substantial benefits when using the \textit{IE as a tool} approach, with a small improvement when tasked to generate a Pandas command. This particularly clear in the larger models (Gemini, GPT, and Claude), which excel on this task, where Claude outperforms the rest of the models. This is evident from the consistent performance improvement across different models when comparing the \textit{IE as a tool} setup with the Generic baseline and other approaches. On the other hand, smaller models showed more moderate improvements when using \textit{IE as a tool}. The gap between the performance in \textit{IE as a tool} and the Gold variant implies a significant potential for enhancing the overall task effectiveness through the refinement of IE tools. Specifically, the larger gap for smaller models suggests that their capabilities can be dramatically increased by improving the accuracy and reliability of the generated tables and commands. This points towards the critical importance of future work on optimizing IE tools to maximize the end-task performance for complex reasoning tasks.

\section{Performance Analysis via \benchmarkname Decomposition}
\label{sec:decomposed_exp}
To understand the factors contributing to the suboptimal performance of current LLMs on \benchmarkname, we decompose the problem into two constituent tasks: 1) table generation from text based on a \benchmarkname instruction and the corresponding text (Section \ref{subsec:table_gen}), and 2) Pandas query generation based on the corresponding table (either gold or previously generated), instruction, and text (Section \ref{subsec:pandas_gen}). Successful execution of these two tasks would naturally result in accurate \benchmarkname results. We assess current LLM capabilities on each task using gold outputs, revealing substantial headroom and a potential for improvement. This observation directly motivated our design of \textit{IE as a Tool}, a method that demonstrably improved TACT performance.

\subsection{Evaluating the Accuracy of Table Generation}
\label{subsec:table_gen}

We assess the capabilities of LLMs to generate the appropriate tables given a \benchmarkname instruction and its corresponding text, tasking the model to construct the correct table based on the specified instruction. Note that the model should infer the correct table from the \benchmarkname instruction, which only implicitly points towards the relevant information to extract. Each model is provided with \benchmarkname instructions and corresponding texts. The task requires generating tables that accurately reflect the data described in the text, and helps to seek the correct information given the instruction. 

We follow the evaluation protocol from \citet{jiao-etal-2023-instruct} and adopt a soft matching strategy \citep{jiao-etal-2022-open} by using SentenceT5-Large \citep{ni-etal-2022-sentence} to calculate the cosine similarity (multiplied by 100) as the semantic similarity score between the generated table and the gold table, as table contents reflect the quality of extraction. Additionally, we use the ROUGE-L F1 score \citep{lin-2004-rouge} to evaluate the lexical similarity of the generated table to the gold one. We also report the Table Validity rate, where we were able to parse a syntactically correct CSV table from the generated content.

\captionsetup[table]{position=below}
\renewcommand{\arraystretch}{1.1} 
\setlength{\tabcolsep}{5pt} 
\begin{table}[htbp]
    \centering
    \tiny
    \caption{Evaluation Results of Different Models on TACT table generation, measuring semantic similarity, ROUGE-L F1 (lexical matching), and table validity between the generated tables and the gold tables. The best-performing results are highlighted in \textbf{bold}.}
    \resizebox{\textwidth}{!}{
    \begin{tabular}{lccccccccc}
        \toprule
        & \multicolumn{3}{c}{\textbf{Semantic Similarity}} & \multicolumn{3}{c}{\textbf{ROUGE-L F1}} & \multicolumn{3}{c}{\textbf{Table Validity Rate (\%)}} \\
        \cmidrule(lr){2-4} \cmidrule(lr){5-7} \cmidrule(lr){8-10}
        Model & 0-shot & 2-shot & 4-shot & 0-shot & 2-shot & 4-shot & 0-shot & 2-shot & 4-shot \\
        \midrule
        Gemma-7b & 68.6 & 69.1 & 69.4 & 6.5 & 6.6 & 7.1 & 0.6 & 23.1 & 25.5 \\
        Mistral-7b & 73.5 & 72.8 & 72.8 & 4.9 & 6.4 & 7.0 & 1.2 & 30.9 & 34.4 \\
        Llama-2-13b & 73.4 & 72.7 & 73.1 & 4.3 & 5.2 & 5.5 & 42.7 & 43.5 & 47.4 \\
        Mixtral 8x7B & 72.3 & 71.7 & 71.8 & 4.5  & 7.1 & 7.2 & 9.5 & 39.9 & 24.3 \\
        Llama-2-70b & 72.3 & 72.4 & 72.7 & 3.9 & 5.1 & 4.9 & 92.5 & 73.4 & 73.8 \\
        Gemini-Pro & 78.4 & 78.2 & 78.3 & 18.8 & 21.0 & 22.9 & 81.5 & 90.2 & 93.3 \\
        Gemini-Ultra & \textbf{78.6} & 78.6 & 79.3 & 18.7 & 21.1 &24.8 & 81.3 & 89.9 & 94.1 \\
        GPT-4o & 78.2 & \textbf{78.9} & 79.9 & 19.3 & \textbf{23.2} & 27.3 & 93.4 & 93.6 & 95.7 \\
        Claude 3.5 Sonnet & 78.5 & 78.6 & \textbf{80.1} & \textbf{19.7} & 23.1 &\textbf{28.1} & \textbf{94.1} & \textbf{94.2} & \textbf{96.2} \\

        \bottomrule
    \end{tabular}
    }
    \label{tab:tab_generation}
\end{table}

The evaluation of various LLMs on their ability to generate accurate tables based on \benchmarkname instructions is shown in Table \ref{tab:tab_generation}. Claude and GPT mostly outperform other models across all metrics. The Gemini models also shows strong performance but vary across different shots, indicating potential instability in its output quality. Llama-2-13b, Llama-2-70b, and Mistral-7b exhibit moderate performance, with Mistral-7b achieving a higher semantic similarity but lower table validity rates. Gemma-7b and Mixtral 8x7B show comparatively lower performance, particularly in table validity. Notably, the smaller models like Gemma-7b and Mistral-7b benefit significantly from few-shot learning, demonstrating that small models are incapable of solving this task without any aid.

While the semantic similarity between the generated tables of Gemini-Ultra and the gold standard tables is relatively high, lexical similarity remains low. However, a qualitative analysis suggests that the generated tables contain key information that addresses the instructions, despite their differences from the gold tables. This observation supports the use of semantic similarity as a more appropriate metric for evaluating table generation in this context \citep{jiao-etal-2023-instruct}.

\subsection{Evaluating the Accuracy of Pandas Command Generation}
\label{subsec:pandas_gen}
The ability to accurately generate Pandas commands is a key intermediate step in solving \benchmarkname queries, and evaluates how well LLMs can comprehend the \benchmarkname instruction and the table at once. 
Thus, we next evaluate the ability of LLMs to generate Pandas commands, when provided with the \benchmarkname instruction, the associated text, as well as a table extracted from the text.
We consider two cases: one where the provided table is the gold one, and one where it is the one generated by the model.
To assess the quality of the generated Pandas queries, we execute them using a Python interpreter, and compare the output to the gold answer.

\captionsetup[table]{position=below}
\renewcommand{\arraystretch}{1.1} 
\setlength{\tabcolsep}{8pt} 
\begin{table}[htbp]
    \centering
    \tiny
    \caption{Evaluation Results of Different Models on TACT Pandas command generation on the generated/gold table, measured by the accuracy after executing the command with a Python interpreter. The best-performing results are highlighted in \textbf{bold}.}
    \resizebox{\textwidth}{!}{
    \begin{tabular}{lccc}
        \toprule
        \textbf{Model} & \textbf{0-shot (Generated/Gold)} & \textbf{2-shot (Generated/Gold)} & \textbf{4-shot (Generated/Gold)} \\
        \midrule
        Gemma-7b & 0 / 0.4 & 1.4 / 2.3 & 1.9 / 2.4 \\
        Mistral-7b & 0.0 / 0.1 & 0.4 / 0.6 & 0.5 / 0.9 \\
        Llama-2-13b & 0.0 / 0.3 & 0.1 / 0.6 & 0.3 / 1.2 \\
        Mixtral 8x7B & 1.2 / 1.8 & 1.3 / 2.1 & 1.5 / 2.9 \\
        Llama-2-70b & 2.5 / 3.4 & 3.1 / 4.1 & 3.2 / 4.3 \\
        Gemini-Pro & 3.4 / 3.6 & 3.0 / 4.8 & 7.3 / 8.0 \\
        Gemini-Ultra & 1.1 / 1.9 & 4.8 / 5.0 & 7.7 / 8.4 \\
        GPT-4o & 4.5 / 4.9 & 5.3 / 6.0 & 8.7 / 9.4 \\
        Claude 3.5 Sonnet & \textbf{5.1} / \textbf{5.9} & \textbf{5.6} / \textbf{6.4} & \textbf{9.6} / \textbf{10.1} \\
        \bottomrule
    \end{tabular}
    }
    \label{tab:pandas_gen}
\end{table}

Table \ref{tab:pandas_gen} presents the results, where Claude consistently outperforms the other models. GPT, Gemini, and Llama-2-70b also demonstrate relatively strong performance, though with some variability across different shot configurations. Interestingly, as in the previous experiment, the smaller models—such as Gemma-7b and Mistral-7b—showed lower performance overall but exhibited significant improvements with few-shot learning, highlighting their ability to effectively leverage additional examples. Llama-2-13b and Mixtral 8x7B delivered moderate performance but still trailed behind the larger models.

It is worth noting that the overall numbers in Table \ref{tab:pandas_gen} are quite low, even when compared to the results in Table \ref{tab:tact_res}, including for gold-standard tables usage. This may seem surprising at first, but upon inspection, we found that many of the Pandas commands generated by the models contain syntax errors or other issues that lead to execution failures. In contrast, the results in Table \ref{tab:tact_res} do not involve Python execution, which allows for more robust command interpretation and, as a result, better answers.

\section{Related Work}
\label{sec:related}

\paragraph{Information Extraction (IE) and Text-to-Table} IE is the process of automatically extracting structured information from unstructured text, involving sub-tasks like named entity recognition, relation extraction, and event extraction. Many works have leveraged large language models (LLMs) to provide effective solutions for IE \citep{ma-etal-2023-large, lu-etal-2023-event, zhou-et-al-2024}. Recently, \citet{wu-etal-2022} presented the concept of text-to-table, and \citet{jiao-etal-2023-instruct} introduced InstructIE, a benchmark that includes triplets of an IE instruction, their associated text, and the relevant content in a tabular format (see Section \ref{sec:dataset}). We employ a labeling methodology on top of InstructIE to distill an aggregative instruction following challenge set. \citet{yuan-etal-2024} presented an effort with a similar focus on numerical tasks but with a narrower scope, limited to financial data. Another related research realm is open information extraction, which aims to extract information without predefined schemas, typically focusing on simple structures from short texts \citep{banko-et-al-2007, mausam-etal-2012-open, stanovsky-etal-2018-supervised, zhan-zhao-etal-2020}.

\paragraph{Complex and Numerical Reasoning} A persistent challenge for LLMs lies in their ability to solve numerical problems, particularly those involving mathematical calculations \citep{geva-etal-2020-injecting, imani-etal-2023, chang-etal-survey-2023, ahn-etal-2024} or the need to stitch and aggregate information across the text \citep{li2023loogle,sprague2024musr,jacovi2024coverbench}. Some works propose to evaluate models on such tasks, by presenting challenge datasets based on financial data  \citep{chen-etal-fin-2021, yuan-etal-2024}. DROP \citep{dua-etal-drop-2019} and IIRC \citep{ferguson-etal-iirc-2020}, two reading comprehension benchmarks involving reasoning, showcase the complexity of comprehensive numerical reasoning. DROP focuses on discrete reasoning over paragraphs, requiring models to perform operations like addition, counting, and sorting, while IIRC evaluates the ability to handle incomplete contexts and locate additional sources of information. \benchmarkname focuses on a more practical and common use-case: reasoning over texts given natural language instructions, necessitating the integration of information scattered throughout the text.

\paragraph{Semantic Parsing} is the process of converting natural language into a machine-interpretable representation, such as a formal query or command \citep{pasupat-liang-2015-compositional,yoran_etal_2022,mekala-etal-2023-zerotop,bogin_etal_2024}. \textit{IE as a tool} also aligns with previous research on semantic parsing, as we utilize executable Pandas command generation over texts and tables, demonstrating how LLMs can interpret and convert complex instructions into executable code operations. 

\paragraph{Multi-step Reasoning and LLM Tools}
Our research is closely related to various methodologies that utilize LLM tools for task resolution, as explored in recent studies \citep{parisi2022talm,mialon2023augmented,schick2023toolformer,hao2023toolkengpt,patil2023gorilla}. These methods train LLMs to utilize APIs independently during inference, contrasting with our \textit{IE as a tool} approach, which employs a static strategy for addressing complex reasoning tasks. Furthermore, prior research has emphasized enhancing task resolution through multi-step processes \citep{berant-etal-2014-modeling,drozdov2023compositional,zhou2023leasttomost,fu2023complexitybased}. Unlike these approaches, which apply general multi-step reasoning or tool triggering across various domains, \textit{IE as a tool} specifically concentrates on constructing tables and executing commands for numerical reasoning, thereby targeting a more focused application of multi-step numerical reasoning.
\section{Conclusion}
\label{sec:conclusion}
In this paper, we introduced \benchmarkname---Text And Calculations through Tables, a dataset designed to assess the reasoning capabilities of Large Language Models (LLMs) through complex, aggregative instructions. \benchmarkname features numerical instructions that require processing and integrating information dispersed across one or more texts for producing the correct answer. By leveraging the InstructIE dataset \citep{jiao-etal-2023-instruct}, experts annotated and transformed instances into a format suitable for aggregative instruction following, ensuring high precision and relevance. To better understand the performance of LLMs on \benchmarkname, we provide further analysis that evaluates performance on two distinct sub-tasks that are likely to be relevant for solving \benchmarkname (table-generation and Pandas command-generation). We also provide a modeling scheme, \textit{IE as a tool}, that is based on this decomposition, and show that it improves performance on \benchmarkname.
Future work could focus on further enhancing the performance of LLMs on \benchmarkname by developing and integrating new, more sophisticated tools that are specifically designed for handling complex, aggregative instructions over text.
\section*{Acknowledgements}
\label{sec:acknowledgements}

We thank Jonathan Berant, Yonatan Bitton, Mor Geva, and Eran Ofek for their valuable feedback and constructive suggestions, and Ayelet Shasha Evron for her assistance in designing the figures for this paper.

\bibliographystyle{abbrvnat}
\bibliography{custom}
\clearpage

\appendix

\section{Limitations}
\label{sec:limitations}
While our work and the introduced tools specifically target numerical complex reasoning tasks, they are not designed to address the full spectrum of natural language processing challenges. Consequently, their application to non-numerical tasks may not yield optimal results. Another significant limitation is the sequential use of tools, which can introduce and propagate errors through the processing stages, potentially compromising the accuracy and reliability of the final response. This propagation of errors underscores the need for careful handling and validation at each step to mitigate compounding inaccuracies. Future work should aim to develop more versatile tools and methodologies that can handle a broader range of tasks while minimizing the risk of error propagation.

\section{License and Intended Use}
\label{sec:license}
The \benchmarkname benchmark, including templates and instructions, is licensed under the Creative Commons Attribution NoDerivs 4.0 International License (CC BY ND 4.0). Contributions derived from InstructIE \citep{jiao-etal-2023-instruct}, such as texts and tables, are provided under the terms of this license. Users must assume responsibility for their use in accordance with the obligations to the creators of InstructIE. The intended use of these materials is for the improvement and evaluation of large language models (LLMs), and we assume full responsibility for any potential violations of rights and confirm adherence to the licensing agreements associated with the data used in this study.\footnote{The data, the full licence details are all available in \url{https://huggingface.co/datasets/google/TACT}, and the \benchmarkname metadata is available in \url{https://huggingface.co/api/datasets/google/TACT/croissant}.}

We emphasize the strict use of the \benchmarkname dataset exclusively for evaluation purposes, prohibiting its inclusion in NLP model training datasets to mitigate potential biases and contamination. We implement measures to prevent data contamination as outlined by \citet{jacovi-etal-2023-stop}, and we require that any future redistribution or use of the data adheres to these same guidelines. Additionally, redistribution of any part of the dataset is advised against without robust measures to block web-crawler access. To facilitate the tracing and management of potential data contamination within web-crawled corpora, a distinct 64-character identifier string is appended to each dataset instance.

\section{\benchmarkname Data Creation and Guidelines}
\label{sec:appendix}
In this section, we include additional details and material regarding the data creation process. We provide the official guidelines that were given to the experts for creating the data (Appendix \ref{sec:guidelines}) along with the concluded data creation process (Appendix \ref{sec:process}).

\subsection{Guidelines}
\label{sec:guidelines}
\begin{figure}[htbp]
    \centering
    \includegraphics[width=.92\textwidth]{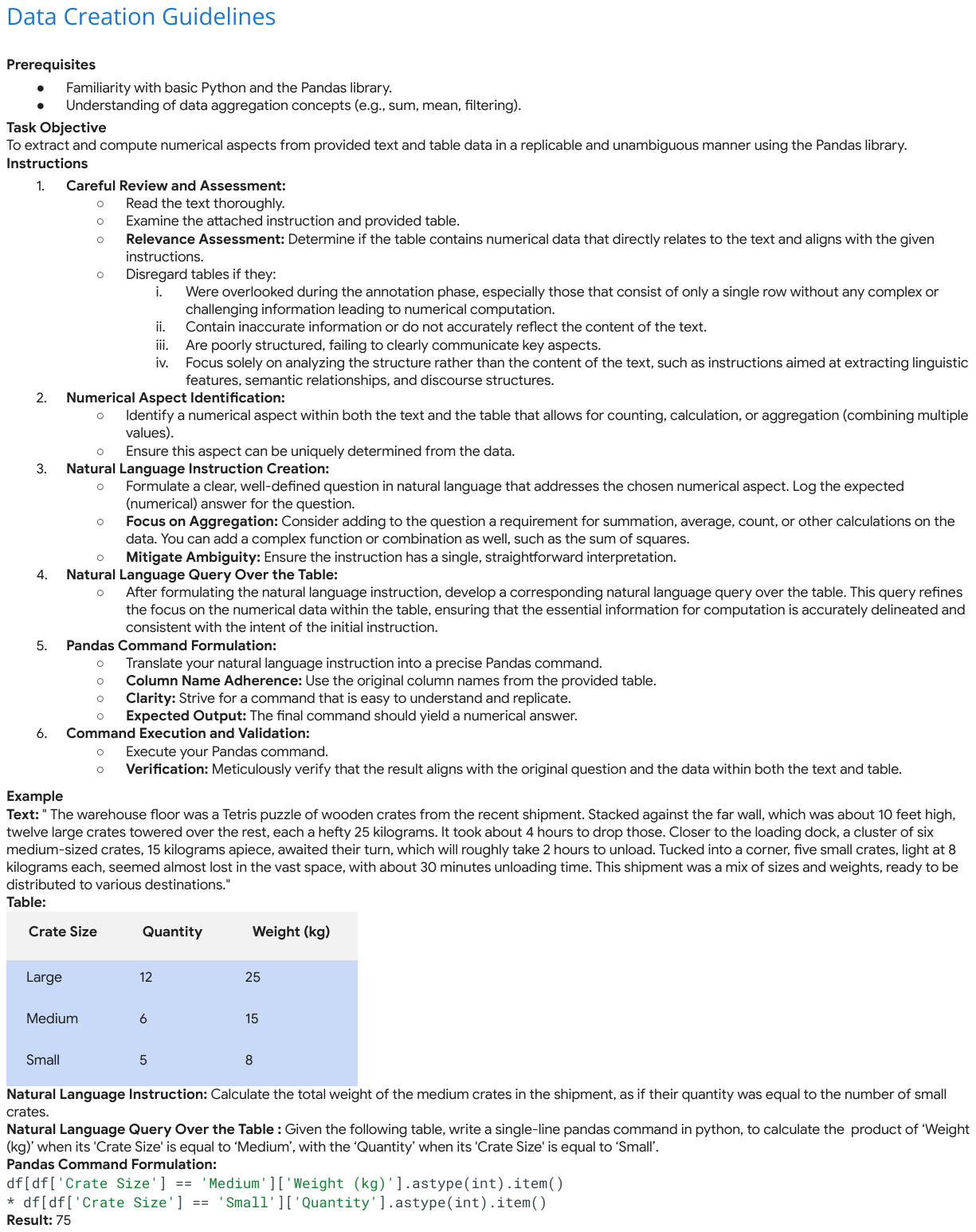}\\
    \caption{The data creation guidelines for the TACT Dataset. This figure presents a comprehensive set of guidelines designed to assist annotators in extracting and computing numerical aspects from provided text and table data using the Pandas library. The guidelines include steps for reviewing and assessing the relevance of data, identifying numerical aspects, formulating natural language instructions and queries, translating these into precise Pandas commands, and validating the results. An example is provided to demonstrate the process, from text and table review to the execution and verification of the computed result.}
    \label{fig:guidelines}
\end{figure}

The data creation process was guided by a comprehensive set of guidelines, prepared to equip the participants with the necessary skills and knowledge for the task. These prerequisites included familiarity with basic Python, proficiency in the Pandas library, and an understanding of data aggregation concepts such as sum, mean, and filtering. The original instructions are depicted in Figure \ref{fig:guidelines}.

\subsection{Creation Process}
\label{sec:process}
\begin{figure}[htbp]
    \centering
    \includegraphics[width=.99\textwidth]{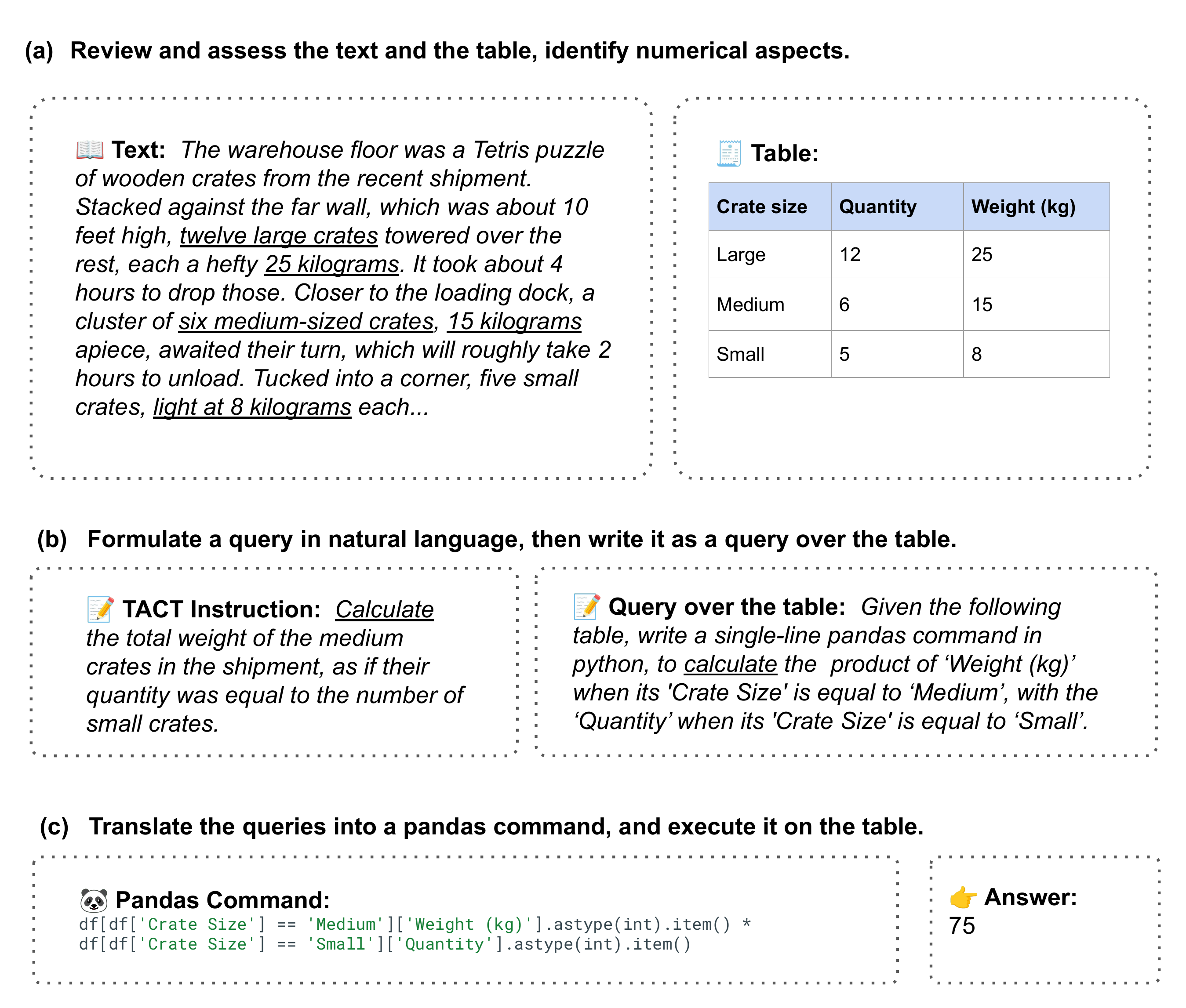}\\
    \caption{The summarized data creation process for the \benchmarkname benchmark. This illustration outlines the systematic guidelines employed for annotating numerical data derived from textual and tabular content within the TACT framework. It elaborates on the sequential steps necessary for annotators to effectively review text and tables, identify numerical data, formulate and translate these into natural language instructions and corresponding Pandas queries, and finally, execute and validate these commands. An example accompanies the instructions to showcase the entire process from initial review to the successful execution and verification of the result.}
    \label{fig:process}
\end{figure}

\begin{figure}[htbp]
    \centering
    \includegraphics[width=1\textwidth]{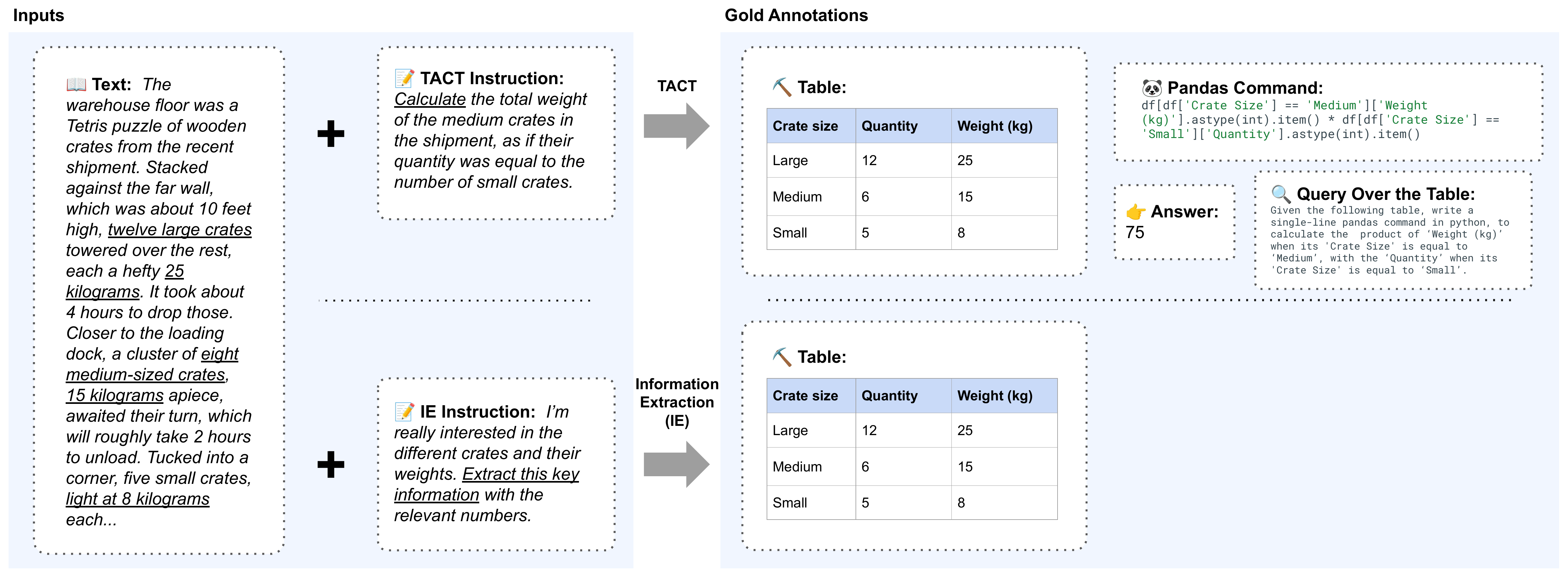}\\
    \caption{\benchmarkname (above) and typical Information Extraction (IE) (below). \benchmarkname includes an instruction, and allows various intermediate outputs before the answer generation, while IE focuses on table generation. The \benchmarkname instruction, the Pandas command, the query over the table, and the answer, were all created by NLP experts.}
    \label{fig:tact_vs_instructie_both}
\end{figure}

In the creation of \benchmarkname, we engaged NLP and data science experts (with a PhD degree specializing in NLP), each with a minimum of four years of experience, to ensure high-quality data curation. These experts reported that labeling each example typically required between 16 to 20 minutes. Approximately 70\% of this time was dedicated to carefully reading the provided tables and accompanying texts, and formulating challenging numerical questions that draw on the data. The remaining time was allocated to writing the queries, including one in Pandas and two in natural language, and executing the Pandas query on the table to verify that the intended results were achieved. This meticulous process guarantees that our dataset can both challenge LLMs and accurately reflects realistic scenarios where mathematical reasoning is essential. The summarized data creation process with an accompanying example, following the guidelines above and the description in Section \ref{subsec:tact}, is illustrated in Figure \ref{fig:process}. The resulting dataset and its components are compared to InstructIE in Figure \ref{fig:tact_vs_instructie_both}. \citep{jiao-etal-2023-instruct}

\section{Additional Experimental Details}
\label{sec:appendix_exp}
In this section, we include additional experimental details, including the Large Language Models' (LLMs') settings we used in our experiments (Appendix \ref{sec:setting}), the few-shot prompting setup (Appendix \ref{sec:appendix_few_shots}), as well as the templates we used (Appendix \ref{sec:appendix_prompts}).

\subsection{Models and Inference Details}
\label{sec:setting}
In our experiments, all large language models (LLMs) were configured to operate with a temperature setting of 0.8. This parameter choice was aimed at striking an optimal balance between diversity and coherence in the model's responses, facilitating more varied yet still plausible outputs for the generation of the final answer as well as the tools. For tasks involving table generation and Pandas command generation, we parse the model's response to extract the desired output from the generated text. Conversely, for tasks requiring a numerical answer, we identify and consider the final number in the generated text as the definitive response.

For the inference of open-source models in our study -- Llama-2-13b-chat\footnote{\url{huggingface.co/meta-llama/Llama-2-13b-chat-hf}}, Llama-2-70b-chat\footnote{\url{huggingface.co/meta-llama/Llama-2-70b-chat-hf}} \citep{touvron2023llama}, Gemma-7b-it\footnote{\url{huggingface.co/google/gemma-1.1-7b-it}} \citep{team2024gemma}, Mistral-7b-instruct\footnote{\url{huggingface.co/mistralai/Mistral-7B-Instruct-v0.2}} \citep{jiang2023mistral}, and Mixtral 8x7B\footnote{\url{huggingface.co/mistralai/Mixtral-8x7B-Instruct-v0.1}} \citep{jiang2024mixtral}, we utilized 8 H100 GPUs. For the rest of the evaluated models, Gemini-1.0-Ultra, Gemini-1.0-Pro \citep{team2023gemini}, we used the developers API\footnote{\url{ai.google.dev}}.

\subsection{Few-shot Prompting}
\label{sec:appendix_few_shots}
In the few-shot evaluation settings, prompts were constructed by randomly choosing few-shot examples out of a pool of 4 demonstrations (that were manually curated from the validation set of InstructIE). These prompts were adjusted to include the maximum number of demonstrations that, along with the query, stayed within the context length limit for all the evaluated models. We ended up with 4 different prompts for each example, to maintain uniformity in the number of shots. For zero-shot examples we ran the same prompt, but the temperature sampling maintained randomness. This approach resulted in effectively 496 examples for each performance result report in Sections \ref{sec:tact_exp_and_setup} and \ref{sec:decomposed_exp}. The robustness of this methodology allowed us to achieve statistically significant results, demonstrating the reliability and efficacy of our few-shot evaluation framework in mimicking real-world analytical tasks.

\subsection{Prompt Templates}
\label{sec:appendix_prompts}
Next, we detail the template structures employed for our few-shot prompts. It's important to note that we tested multiple templates and settings to ensure that substantial effort was devoted to reporting the strongest baseline prompt implementations. Our approach included the creation of prompts designed for the specific tasks in this paper, each complemented by examples that demonstrate their goal. The Generic \benchmarkname task prompt is depicted in Figure \ref{fig:generic_prompt}, and the Table Generation task prompt, Table Generation task prompt, and the overall task prompt are illustrated in Figures \ref{fig:tab_gen_prompt}, \ref{fig:pandas_gen_prompt}, and \ref{fig:tact_prompt}, respectively. Figure \ref{fig:tact_prompt} matches both the \textit{In-context IE} and \textit{IE as a tool} setups, as for the latter we insert the generated table and Pandas command as the outputs from the tools in Figures \ref{fig:tab_gen_prompt} and \ref{fig:pandas_gen_prompt} (see Section \ref{sec:tact_exp_and_setup} for the setups' details). Note that for the Chain-of-Thought (CoT) setups, the ``Let's think step-by-step'' suffix was added. 

\captionsetup[figure]{skip=12pt} 
\begin{figure}[t]

\lstdefinestyle{promptStyle}
{
    basicstyle={\footnotesize\ttfamily},
    numbers=left,numberstyle=\footnotesize,
    xleftmargin=2.8em,
    xrightmargin=1.5em,
    showstringspaces=false,
      showspaces=false,
        showtabs=false,
    tabsize=2,
    breaklines=true,
        flexiblecolumns=true,
        escapeinside={<@}{@>},
          breakatwhitespace=true
}

\newtcblisting{mylisting}[1]{
  enhanced,
  listing only,
  boxrule=0.8pt,
  sharp corners=downhill,
  top=0mm,
  bottom=0mm,
  left=2mm,
  right=0mm,
  boxsep=0mm,
  colframe=black,
  colback=white,
  listing options={
    style=#1
  }
}

\definecolor{instructionsColor}{rgb}{0.1, 0.5, 0.1}

\begin{mylisting}{promptStyle}
<@\textcolor{instructionsColor}{In this task, you should calculate the numerical answer that is the solution for the given instruction and text. The answer should be numerical, using digits only.}@>

<@\color{red}Example1:@>
<@\color{blue}Instruction:@> Calculate the sum of the year that "To Kill a Mockingbird" was published in and the year that it won a prize according to the text.
<@\color{blue}Text:@> ... It was published in 1960 ... A year after its release, it won the Pulitzer Prize ...
<@\color{blue}Answer:@> 4021

<@\color{red}Example2:@>
<@\color{blue}Instruction:@> Count the number of achievements that include instructions.
<@\color{blue}Text:@> 1. We present FLAIR ... 2. How well can NLP models perform? ... 3. Pretrained language models have become increasingly prominent ...
<@\color{blue}Answer:@> 3

<@\color{red}Example3:@>
<@\color{blue}Instruction:@> Calculate the sum of squares of the stock price increases in the text.
<@\color{blue}Text:@> ... The S&P 500 rose 1.45% ... the Nasdaq Composite popped 1.07% ... The Dow Jones Industrial Average led gains, rising 2.12% ...
<@\color{blue}Answer:@> 0.0558

<@\textcolor{red}{Now your turn:}@>
<@\color{blue}Instruction:@> [Your custom instruction here]
<@\color{blue}Text:@> [Your relevant text spans here]
<@\color{blue}Answer:@>
...
\end{mylisting}
\caption{Example of the Generic \benchmarkname task prompt provided to our evaluated models. The prompt consists of basic instructions, several few-shot in-context examples, and the instance input. The examples demonstrate how to generate the answer to the instruction, based on given instruction and text. Note that the examples illustrated are slightly modified from the original \benchmarkname, for simplicity.}
\label{fig:generic_prompt}
\end{figure}
\captionsetup[figure]{skip=12pt} 
\begin{figure}[t]

\lstdefinestyle{promptStyle}
{
    basicstyle={\footnotesize\ttfamily},
    numbers=left,numberstyle=\footnotesize,
    xleftmargin=2.8em,
    xrightmargin=1.5em,
    showstringspaces=false,
      showspaces=false,
        showtabs=false,
    tabsize=2,
    breaklines=true,
        flexiblecolumns=true,
        escapeinside={<@}{@>},
          breakatwhitespace=true
}

\newtcblisting{mylisting}[1]{
  enhanced,
  listing only,
  boxrule=0.8pt,
  sharp corners=downhill,
  top=0mm,
  bottom=0mm,
  left=2mm,
  right=0mm,
  boxsep=0mm,
  colframe=black,
  colback=white,
  listing options={
    style=#1
  }
}

\definecolor{instructionsColor}{rgb}{0.1, 0.5, 0.1}

\begin{mylisting}{promptStyle}
<@\textcolor{instructionsColor}{In this task, you should create an appropriate CSV table according to the given Instruction and Text.}@>

<@\color{red}Example1:@>
<@\color{blue}Instruction:@> Calculate the sum of the year that "To Kill a Mockingbird" was published in and the year that it won a prize according to the text.
<@\color{blue}Text:@> ... It was published in 1960 ... A year after its release, it won the Pulitzer Prize ...
<@\color{blue}Table:@>
"Event","Year"
"Published","1960"
"Prize Won","1961"


<@\color{red}Example2:@>
<@\color{blue}Instruction:@> Count the number of achievements that include instructions.
<@\color{blue}Text:@> 1. We present FLAIR ... 2. How well can NLP models perform? ... 3. Pretrained language models have become increasingly prominent ...
<@\color{blue}Table:@>
"Number","Achievement"
"1","We present FLAIR"
"2","How well can NLP models perform?"
"3","Pretrained language models prominence"


<@\color{red}Example3:@>
<@\color{blue}Instruction:@> Calculate the sum of squares of the stock price increases in the text.
<@\color{blue}Text:@> ... The S&P 500 rose 1.45% ... the Nasdaq Composite popped 1.07% ... The Dow Jones Industrial Average led gains, rising 2.12% ...
<@\color{blue}Table:@>
"Increase","Venture"
"1.45%","S&P"
"1.07%","Nasdaq Composite popped"
"2.12%","Dow Jones Industrial"


<@\textcolor{red}{Now your turn:}@>
<@\color{blue}Instruction:@> [Your custom instruction here]
<@\color{blue}Text:@> [Your relevant text spans here]
<@\color{blue}Table:@>
...
\end{mylisting}
\caption{Example table generation prompt provided to our evaluated models. The prompt consists of basic instructions, several few-shot in-context examples, and the instance input. The examples demonstrate how to create a CSV table based on given instructions and text. Note that the examples illustrated are slightly modified from the original \benchmarkname, for simplicity.}
\label{fig:tab_gen_prompt}
\end{figure}

\captionsetup[figure]{skip=12pt} 
\begin{figure}[t]

\lstdefinestyle{promptStyle}
{
    basicstyle={\footnotesize\ttfamily},
    numbers=left,numberstyle=\footnotesize,
    xleftmargin=2.8em,
    xrightmargin=1.5em,
    showstringspaces=false,
      showspaces=false,
        showtabs=false,
    tabsize=2,
    breaklines=true,
        flexiblecolumns=true,
        escapeinside={<@}{@>},
          breakatwhitespace=true
}

\newtcblisting{mylisting}[1]{
  enhanced,
  listing only,
  boxrule=0.8pt,
  sharp corners=downhill,
  top=0mm,
  bottom=0mm,
  left=2mm,
  right=0mm,
  boxsep=0mm,
  colframe=black,
  colback=white,
  listing options={
    style=#1
  }
}

\definecolor{instructionsColor}{rgb}{0.1, 0.5, 0.1}

\begin{mylisting}{promptStyle}
<@\textcolor{instructionsColor}{In this task, you should create an appropriate Pandas command according to the given Instruction, Text, and Table, such that the command will run on the table and return the correct number answering the instruction.}@>

<@\color{red}Example1:@>
<@\color{blue}Instruction:@> Calculate the sum of the year that "To Kill a Mockingbird" was published in and the year that it won a prize according to the text.
<@\color{blue}Text:@> ... It was published in 1960 ... A year after its release, it won the Pulitzer Prize ...
<@\color{blue}Table:@>
"Event","Year"
"Published","1960"
"Prize Won","1961"
<@\color{blue}Pandas Command:@>
df['Year'].astype(int).sum()


<@\color{red}Example2:@>
<@\color{blue}Instruction:@> Count the number of achievements that include instructions.
<@\color{blue}Text:@> 1. We present FLAIR ... 2. How well can NLP models perform? ... 3. Pretrained language models have become increasingly prominent ...
<@\color{blue}Table:@>
"Number","Achievement"
"1","We present FLAIR"
"2","How well can NLP models perform?"
"3","Pretrained language models prominence"
<@\color{blue}Pandas Command:@>
len(df)


<@\color{red}Example3:@>
<@\color{blue}Instruction:@> Calculate the sum of squares of the stock price increases in the text.
<@\color{blue}Text:@> ... The S&P 500 rose 1.45% ... the Nasdaq Composite popped 1.07% ... The Dow Jones Industrial Average led gains, rising 2.12% ...
<@\color{blue}Table:@>
"Increase","Venture"
"1.45%","S&P"
"1.07%","Nasdaq Composite popped"
"2.12%","Dow Jones Industrial"
<@\color{blue}Pandas Command:@>
(df['Increase'].str.replace('%', '').astype(float) / 100).pow(2).sum()


<@\textcolor{red}{Now your turn:}@>
<@\color{blue}Instruction:@> [Your custom instruction here]
<@\color{blue}Text:@> [Your relevant text spans here]
<@\color{blue}Table:@> [Your relevant table here]
<@\color{blue}Pandas Command:@>
...
\end{mylisting}
\caption{Example Pandas command generation prompt provided to our evaluated models. The prompt consists of basic instructions, several few-shot in-context examples, and the instance input. The examples demonstrate how to create a Pandas command based on given instructions, text, and table. Note that the examples illustrated are slightly modified from the original \benchmarkname, for simplicity.}
\label{fig:pandas_gen_prompt}
\end{figure}

\captionsetup[figure]{skip=12pt} 
\begin{figure}[t]

\lstdefinestyle{promptStyle}
{
    basicstyle={\footnotesize\ttfamily},
    numbers=left,numberstyle=\footnotesize,
    xleftmargin=2.8em,
    xrightmargin=1.5em,
    showstringspaces=false,
      showspaces=false,
        showtabs=false,
    tabsize=2,
    breaklines=true,
        flexiblecolumns=true,
        escapeinside={<@}{@>},
          breakatwhitespace=true
}

\newtcblisting{mylisting}[1]{
  enhanced,
  listing only,
  boxrule=0.8pt,
  sharp corners=downhill,
  top=0mm,
  bottom=0mm,
  left=2mm,
  right=0mm,
  boxsep=0mm,
  colframe=black,
  colback=white,
  listing options={
    style=#1
  }
}

\definecolor{instructionsColor}{rgb}{0.1, 0.5, 0.1}

\begin{mylisting}{promptStyle}
<@\textcolor{instructionsColor}{In this task, you should calculate the numerical answer that is the solution for the given instruction and text. Use the following Table in your calculations, by executing the Pandas Command on it. The answer should be numerical, using digits only.}@>

<@\color{red}Example1:@>
<@\color{blue}Instruction:@> Calculate the sum of the year that "To Kill a Mockingbird" was published in and the year that it won a prize according to the text.
<@\color{blue}Text:@> ... It was published in 1960 ... A year after its release, it won the Pulitzer Prize ...
<@\color{blue}Table:@>
"Event","Year"
"Published","1960"
"Prize Won","1961"
<@\color{blue}Pandas Command:@>
df['Year'].astype(int).sum()
<@\color{blue}Answer:@> 4021

<@\color{red}Example2:@>
<@\color{blue}Instruction:@> Count the number of achievements that include instructions.
<@\color{blue}Text:@> 1. We present FLAIR ... 2. How well can NLP models perform? ... 3. Pretrained language models have become increasingly prominent ...
<@\color{blue}Table:@>
"Number","Achievement"
"1","We present FLAIR"
"2","How well can NLP models perform?"
"3","Pretrained language models prominence"
<@\color{blue}Pandas Command:@>
len(df)
<@\color{blue}Answer:@> 3

<@\color{red}Example3:@>
<@\color{blue}Instruction:@> Calculate the sum of squares of the stock price increases in the text.
<@\color{blue}Text:@> ... The S&P 500 rose 1.45% ... the Nasdaq Composite popped 1.07% ... The Dow Jones Industrial Average led gains, rising 2.12% ...
<@\color{blue}Table:@>
"Increase","Venture"
"1.45%","S&P"
"1.07%","Nasdaq Composite popped"
"2.12%","Dow Jones Industrial"
<@\color{blue}Pandas Command:@>
(df['Increase'].str.replace('%', '').astype(float) / 100).pow(2).sum()
<@\color{blue}Answer:@> 0.0558

<@\textcolor{red}{Now your turn:}@>
<@\color{blue}Instruction:@> [Your custom instruction here]
<@\color{blue}Text:@> [Your relevant text spans here]
<@\color{blue}Table:@> [Your relevant table here]
<@\color{blue}Pandas Command:@> [Your Pandas command here]
<@\color{blue}Answer:@>
...
\end{mylisting}
\caption{Example of the \benchmarkname task prompt provided to our evaluated models, using tables and Pandas commands. The prompt consists of basic instructions, several few-shot in-context examples, and the instance input. The examples demonstrate how to generate the answer to the instruction, based on given instruction, text, table, and Pandas command, and should generate the computed answer. Note that the examples illustrated are slightly modified from the original \benchmarkname, for simplicity.}
\label{fig:tact_prompt}
\end{figure}
\end{document}